\documentclass[letterpaper, 10 pt, conference]{ieeeconf}  

\IEEEoverridecommandlockouts                              

\usepackage{amsmath,amsfonts}
\usepackage{bm}
\usepackage[table,xcdraw]{xcolor}
\usepackage{url}
\usepackage{graphicx}
\usepackage{mathtools}
\usepackage{dirtree}
\usepackage{amssymb} 
\usepackage{breqn}

\usepackage{multirow}
\usepackage{cancel} 
\usepackage{xfrac}
\usepackage{xspace}
\usepackage{booktabs}
\usepackage{pifont}
\usepackage{arydshln}

\makeatletter
\let\NAT@parse\undefined
\makeatother
\usepackage[colorlinks=true,
			linkcolor=blue,
			urlcolor=blue,
			citecolor=blue]{hyperref}

\newcommand{\refSec}[1]{Sec.~\ref{#1}} 
\newcommand{\refFig}[1]{Fig.~\ref{#1}}
\newcommand{\refTab}[1]{Table~\ref{#1}}

\newcommand{\boldVar}[1]{
	\bm{#1}
}

\newcommand{\parenthesis}[1]{
	\left(#1\right)
}

\newcommand{\inR}[1]{
	\in\mathbb{R}^{#1}
}

\newcommand{\R}{\mathbf{R}}

\usepackage{balance}
\usepackage{verbatim}
\definecolor{darkgreen}{rgb}{0.0, 0.7, 0.0}

\newcommand{\cmark}{\textcolor{darkgreen}{\ding{51}}}
\newcommand{\xmark}{{\color{red}\ding{55}}}
\def\ourdataset{\textbf{SLAM\&Render}\xspace}

\begin{document}

\title{\LARGE \bf
\ourdataset: A Benchmark for the Intersection Between Neural Rendering, Gaussian Splatting and SLAM \\ \vspace{1.5mm} \large 
}

\author{Samuel Cerezo$^{1}$, Gaetano Meli$^{1,2}$, Tomás Berriel Martins$^{1}$, Kirill Safronov$^{2}$ and Javier Civera$^{1}$%
\thanks{$^{1}$Authors are with the Departamento de Informática e Ingeniería de Sistemas, Universidad de Zaragoza, 50018 Zaragoza, ES. E-mails: {\tt\small \{sacerezo, tberriel, jcivera\}@unizar.es}}%
\thanks{$^{2}$Authors are with the Technology \& Innovation Center, KUKA Deutschland GmbH, 86165 Augsburg, DE. E-mails: {\tt\small \{gaetano.meli, kirill.safronov\}@kuka.com}}}

\maketitle

\begin{abstract}
Models and methods originally developed for Novel View Synthesis and Scene Rendering, such as Neural Radiance Fields (NeRF) and Gaussian Splatting, are increasingly being adopted as representations in Simultaneous Localization and Mapping (SLAM). 
However, existing datasets fail to include the specific challenges of both fields, such as sequential operations and, in many settings, multi-modality in SLAM or generalization across viewpoints and illumination conditions in neural rendering. Additionally, the data are often collected using sensors which are handheld or mounted on drones or mobile robots, which complicates the accurate reproduction of sensor motions.
To bridge these gaps, we introduce \ourdataset, a fully controlled benchmark designed to isolate and evaluate specific challenges arising at the intersection of SLAM, Neural Rendering, and Gaussian Splatting. 
Recorded with a robot manipulator, it uniquely includes 40 sequences with time-synchronized RGB-D images, IMU readings, robot kinematic and ground-truth data. By releasing robot kinematic data, the dataset also enables the assessment of recent integrations of SLAM methods within robotic applications.
The dataset features five setups with consumer and industrial objects under four controlled lighting conditions, each with separate training and test trajectories. All sequences are static with different levels of object rearrangements and occlusions.
Our experimental results, obtained with several baselines from the literature, validate \ourdataset as a relevant benchmark for this emerging research area.
The dataset can be accessed through this link: \href{https://tinyurl.com/SLAMRender}{
https://tinyurl.com/SLAMRender}.


\end{abstract}

\section{Introduction}

Simultaneous Localization and Mapping (SLAM) is a key enabler for autonomous robot navigation and scene understanding.
Its goal is to estimate the robot's pose while building a consistent map using only onboard sensors. Several high-performing multimodal~\cite{campos2021orb} and unimodal~\cite{Teed2021DROIDSLAM,zhang2014loam} methods are available today, although open challenges still remain~\cite{cadena2016past,macario2022comprehensive}.
SLAM differs from alternative approaches to 3D reconstruction (e.g., Structure
from Motion~\cite{schonberger2016structure}) mainly by its online, real-time and multimodal features, making it well-suited for robotic applications. Emerging works~\cite{klingensmith2016armslam,subeta2022multi,li2019slam} show the benefits of combining SLAM with robot kinematic data in the presence of uncertainties in the kinematic model and/or actuators. Moreover, robot kinematics alone does not inherently address shifts in the extrinsic parameters~\cite{Li2024AutoHandEye}, cross-session relocalization and map merging, and high-quality 3D reconstruction, which are paramount for workspace exploration, maintenance and inspection tasks.
 
Novel View Synthesis (NVS), seemingly distant from robot SLAM, has experienced a resurgence due to groundbreaking research in Neural Radiance Fields (NeRFs)~\cite{mildenhall2021nerf}. It has been quickly adopted as a map representation in visual~\cite{sucar2021imap,zhu2022nice} and LiDAR~\cite{deng2023nerf} SLAM pipelines. On the other hand, Gaussian Splatting~\cite{kerbl20233d}, developed as a much faster alternative to NeRFs, has had an even bigger impact in visual SLAM~\cite{matsuki2024gaussian,keetha2024splatam}. 

Motivated by the growing convergence of visual SLAM, Neural Rendering, and Gaussian Splatting, and by the lack of dedicated benchmarks, we introduce \ourdataset. Recorded with a robot manipulator, it ensures accurate reproductions of camera motions, split in separate \textit{train} and \textit{test} trajectories. The dataset provides 40 static sequences with synchronized RGB-D images, IMU readings, robot kinematic data, camera parameters, and ground-truth data.
It comprises five setups with consumer and industrial objects under four controlled lighting conditions, showcasing object rearrangements with varying levels of occlusion.

All data will be made available online under CC-BY 4.0 license at our website. Additional information, software tools, and videos for visual inspection of the dataset will also be included.

\begin{figure}[t]
    \centering
    {\includegraphics[width=\linewidth]{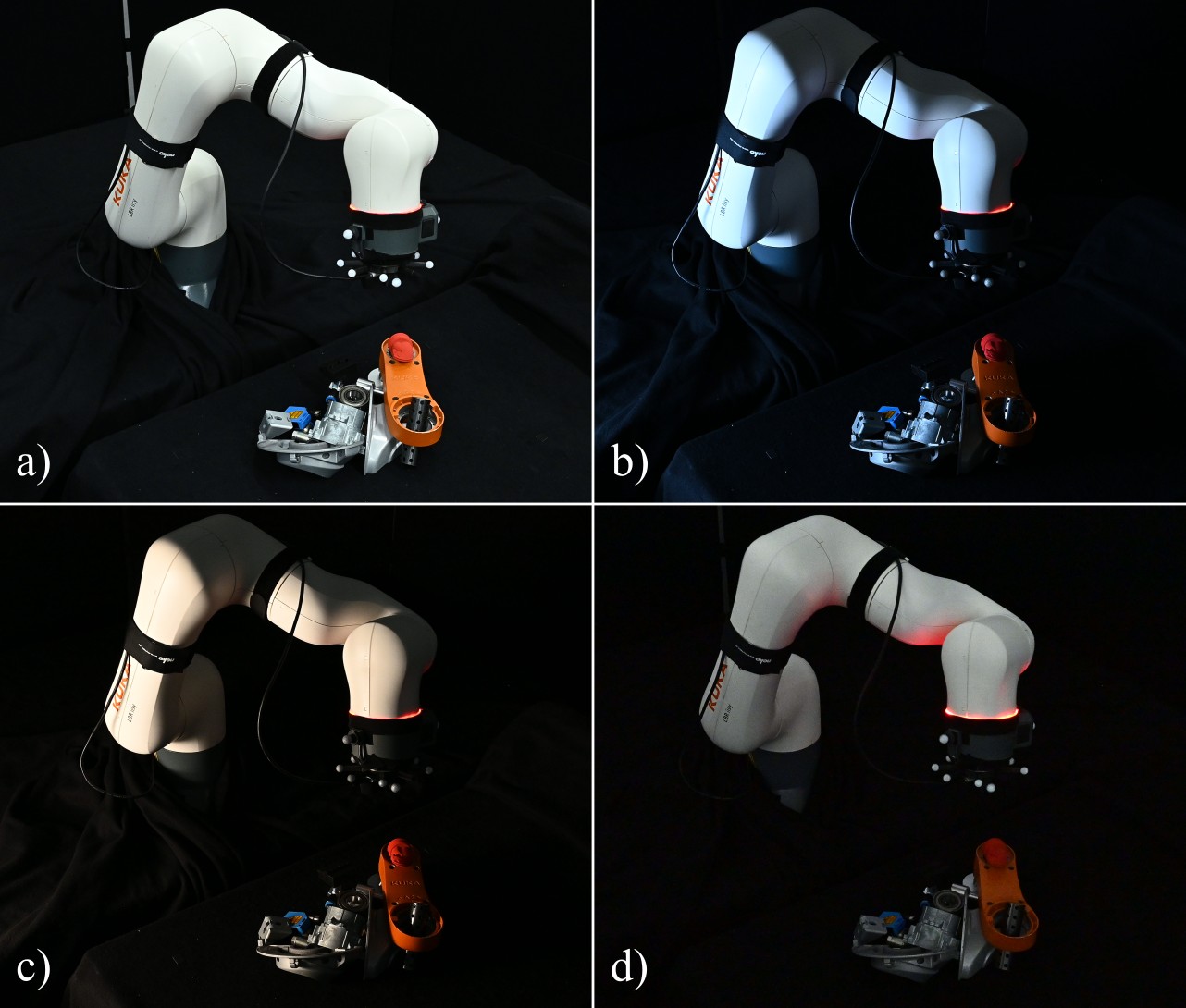}}
    \caption{Illustration of the capture setup for our \ourdataset dataset, recorded from a Intel RealSense at the end effector of a robotic arm that moves around a set of objects on a table. See the four different light conditions present in our dataset: a) natural, b) cold, c) warm, and d) dark}
    \label{fig:light-conditions}
\end{figure}

\section{Related Work}

\begin{table*}[t]
\footnotesize
\centering
\setlength{\tabcolsep}{5pt} 
\rowcolors{2}{gray!15}{white} 
\begin{tabular}{l l c c c c c c c c c}
    Year & Dataset & Use & RGB & Depth & IMU & Enc. & GT & Ill. & Rearr. & TS.\\
    \midrule
    2014 & \href{https://roboimagedata.compute.dtu.dk/?page_id=36}{DTU}~\cite{jensen2014large} & NVS & \cmark & \cmark & \xmark & \xmark & \cmark & \cmark & \xmark & \xmark \\
    2014& \href{https://cirl.lcsr.jhu.edu/research/hmm/datasets/jigsaws_release/}{JIGSAWS}~\cite{gao-jigsaws-dataset} & Motion Model. & \cmark & \xmark & \xmark & \cmark & \xmark & \xmark & \xmark & \xmark  \\
    {2015} & \href{https://www.ycbbenchmarks.com/}{YCB object and Model set}~\cite{calli2015-ycb-dataset} & Robotic Manip. & \cmark & \cmark & \xmark & \xmark & \cmark & \xmark & \xmark & \xmark \\
    2016 & \href{https://projects.asl.ethz.ch/datasets/doku.php?id=kmavvisualinertialdatasets}{EuRoC}~\cite{burri2016euroc} & SLAM & \cmark & \xmark & \cmark & \xmark & \cmark & \cmark & \xmark & \cmark \\
    2017 & \href{https://www.tanksandtemples.org/}{Tanks and Temples}~\cite{Knapitsch2017} & NVS & \cmark & \cmark & \xmark & \xmark & \xmark & \xmark & \xmark & \xmark \\
    2018 & \href{https://cvg.cit.tum.de/data/datasets/visual-inertial-dataset}{TUM VI}~\cite{schubert2018tum} & Odom. & \cmark & \xmark & \cmark & \xmark & \textcolor{darkgreen}{(}\cmark \textcolor{darkgreen}{)} & \xmark & \xmark & \xmark \\
    2019 & \href{https://lifelong-robotic-vision.github.io/dataset/scene.html}{OpenLORIS-Scene}~\cite{shi2020openloris} & SLAM & \cmark & \cmark & \cmark & \cmark & \cmark & \xmark & \xmark & \cmark \\
    2019 & \href{https://github.com/facebookresearch/Replica-Dataset}{Replica}~\cite{Replica-dataset} & SLAM \& NVS & \cmark & \cmark & \cmark & \xmark & \cmark & \xmark & \xmark & \cmark \\
    {2020} & \href{https://ori-drs.github.io/newer-college-dataset/}{Newer College}~\cite{ramezani2020newer} & {SLAM \& NVS} & \cmark & \xmark & \cmark & \xmark & \cmark & \xmark & \xmark & \cmark \\
    2021 & \href{https://jonbarron.info/mipnerf360/}{Mip-NeRF360}~\cite{barron2022mipnerf360} & NVS & \cmark & \xmark & \xmark & \xmark & \xmark & \xmark & \xmark & \xmark \\
    2023 & \href{https://kaldir.vc.in.tum.de/scannetpp/}{ScanNet++}~\cite{yeshwanth2023scannet++} & SLAM \& NVS & \cmark & \cmark & \xmark & \xmark & \cmark & \xmark & \xmark & \cmark \\
    {2025} & \href{https://dynamic.robots.ox.ac.uk/datasets/oxford-spires/}{Oxford Spires}~\cite{tao2025spires} & {SLAM \& NVS} & \cmark & \cmark & \cmark & \xmark & \cmark & \xmark & \xmark & \cmark \\
    \midrule
    2025 & \href{https://anonymous.4open.science/SLAM&Render}{\bf{\ourdataset}} & SLAM \& NVS & \cmark & \cmark & \cmark & \cmark & \cmark & \cmark & \cmark & \cmark\\
    \bottomrule
\end{tabular}
\caption{Overview of commonly used datasets for SLAM and NVS, including available sensor modalities (RGB, Depth, IMU, Encoders), ground-truth camera pose (GT), illumination variations (Ill.), scene rearrangements (Rearr.), and time synchronization (TS).}
\label{tab:datasets}
\end{table*}
Publicly available datasets have played a crucial role in advancing research in SLAM and NVS. \refTab{tab:datasets} provides an overview of the most relevant datasets in this field. Unless otherwise stated, the following datasets \emph{do not} feature (i) robot kinematic data, (ii) full control of lighting conditions, (iii) reproducible sensor motions with high accuracy, and (iv) object rearrangements across sequences.

\subsection{Datasets for SLAM}

The EuRoC dataset~\cite{burri2016euroc} provides visual–inertial data from a micro aerial vehicle organized into two batches: industrial hall sequences for SLAM evaluation, and indoor room sequences for 3D reconstruction. 
Several challenges, such as varying natural
light conditions and dynamic scenes, are included.
The TUM VI dataset~\cite{schubert2018tum} is designed for evaluating odometry algorithms. It contains RGB images, photometric calibration, synchronized IMU data, and ground-truth camera poses, while depth images are not included. 
The OpenLORIS-Scene dataset~\cite{shi2020openloris} consists of visual–inertial, LiDAR and odometry data (i.e., encoder readings) from a mobile robot to benchmark SLAM and scene understanding in everyday scenarios. 
The Newer College~\cite{ramezani2020newer} and the Oxford Spires~\cite{tao2025spires} datasets provide visual-inertial, and LiDAR data from handheld devices. Both include time-synchronized data, calibration parameters for all the involved sensors, loop closures, and cluttered sequences observed from varying viewpoints. 
The Replica dataset~\cite{Replica-dataset} includes 18 photorealistic indoor multi-scale reconstructions with dense meshes, rich textures, and reflective surfaces.
\begin{figure*}[t]
    \centering
    \includegraphics[width=\linewidth]{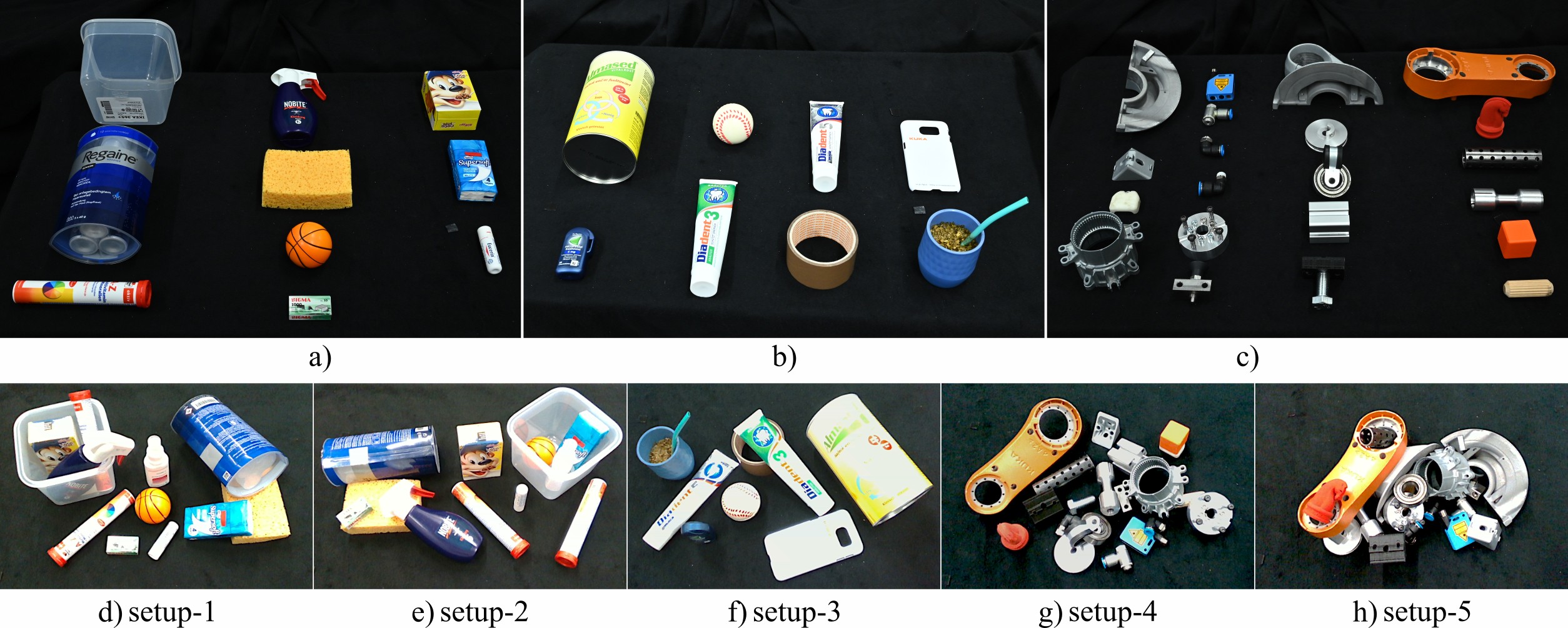}
    \caption{Illustration of the objects included in \ourdataset: a-b) supermarket goods, c) industrial goods, d)~–~h) object arrangements of the five setups.}
\label{fig:objects_and_setups}
\end{figure*}
\subsection{Datasets for Novel View Synthesis}

Despite the popularity of NeRFs, most datasets used for evaluation sample frames along a single camera trajectory, often lacking the diversity and complexity of real-world settings.
DTU~\cite{jensen2014large} was introduced to evaluate multi-view stereo techniques using a robotic setup for object 3D reconstruction tasks. It features systematic light control with a  light scanner, however it addresses a very specific use case.
In the same application field, Tanks and Temples~\cite{Knapitsch2017} was captured in realistic indoor and outdoor settings using an industrial laser scanner for ground-truth collection.
Recently, Mip-NeRF360~\cite{barron2022mipnerf360} introduced a dataset for high-quality NVS. It captured camera trajectories around a central object in indoor and outdoor scenarios with detailed backgrounds.
ScanNet++~\cite{yeshwanth2023scannet++} benchmarks 3D scene understanding methods by combining 3D laser scans, high-quality images, and RGB-D streams with rich semantic annotations and independent camera trajectories for NVS.


Within datasets for robotics applications, the YCB Object and Model Set~\cite{calli2015-ycb-dataset} provides RGB-D images of several object manipulation sequences. In contrast, the JIGSAWS dataset~\cite{gao-jigsaws-dataset} collects time-synchronized stereo images from an endoscopic camera, along with kinematic data from a surgical robot. However, it addresses applications only in the field of surgical activities.


Unlike prior works, our SLAM\&Render uniquely benchmarks SLAM, Neural Rendering, and Gaussian Splatting methods with 40 time-synchronized sequences featuring multimodal data, including RGB-D images, IMU readings, ground-truth, and robot kinematic data. Recorded with a robot manipulator, it allows for accurately reproducing hand-eye camera motions, which are split in train and test trajectories. 
{This is particularly advantageous for ablation studies, as it enables to isolate specific variables (e.g., lighting changes) without considering effects from motion inconsistencies.}
The five sets contain consumer and industrial objects under four controlled lighting conditions, which generally have a negative impact on SLAM and NVS~\cite{ye2024gaussian, martin2021nerf}. {Instead of expanding sensing modalities, SLAM\&Render is, to the best of our knowledge, the first dataset to provide time‑synchronized multi‑sensor and kinematic data, controlled object occlusions, and scene rearrangements within a fully controlled recording setup, \emph{complementing} the existing datasets.}

\section{Notation}




\subsection{Convention for rigid transformations}

{
A rigid transformation between reference frames $A$ and $B$ is defined as 
$\mathbf{T}_{A}^{B}=\left[\mathbf{R}_{A}^{B},\,\mathbf{p}_{A}^{B};\,\mathbf{0}_{1 \times 3},\,1\right]\in SE\parenthesis{3}$, 
where $\R_{A}^{B}\in SO\parenthesis{3}$ and $\mathbf{p}_{A}^{B}\inR{3}$ denote the rotation and translation of $A$ with respect to $B$, respectively. 
For any point $\mathbf{p}^{A}\inR{3}$, its coordinates in $B$ are $\tilde{\mathbf{p}}^{B}=\mathbf{T}_{A}^{B}\tilde{\mathbf{p}}^{A}$, where $\tilde{\mathbf{p}}^{A}$ and $\tilde{\mathbf{p}}^{B}$ are the homogeneous vectors of $\mathbf{p}^{A}$ and $\mathbf{p}^{B}$, respectively.}


\subsection{Forward kinematics}
A manipulator consists of a sequence of rigid bodies (\textit{links}) connected through kinematic \textit{joints}. Let $\boldVar{x}\parenthesis{t}\inR{m}$ be the task variable to be controlled (e.g., end-effector pose) and $\boldVar{q}\parenthesis{t}\inR{n}$ the vector of the variables describing the configuration (i.e, joint positions) of a given $n$-DOFs robot, with $n \geq m$. The relation between $\boldVar{q}$ and $\boldVar{x}$ is given by the forward kinematics
equation
\begin{equation}
    \label{eq:fk}
    \boldVar{x}\parenthesis{t} = {f}\parenthesis{\boldVar{q}\parenthesis{t}},
\end{equation}
where ${f}\parenthesis{\cdot}$ is, in general, a nonlinear function that relates the task variables with the robot configuration.

\section{The SLAM\&Render Dataset}

The proposed dataset comprises 40 {static} sequences that include both inertial and RGB-D data. These were recorded using an Intel RealSense D435i camera, mounted in an eye-to-hand configuration on a KUKA LBR iisy 3 R760 manipulator. The robot has been employed as a precise and reliable sensor carrier, ensuring consistent and reproducible camera trajectories. Additionally, kinematic data were collected from the manipulator. In particular, the dataset contains the robot configurations $\boldVar{q}\parenthesis{t}$ along the motion and the corresponding flange poses computed using \eqref{eq:fk}.
An external Motion Capture System (MCS) was used to obtain the ground-truth camera poses. All sequences were recorded in an environment where the objects were placed on a black background, surrounded by black panels featuring pictures that display complex and colorful patterns (see \refSec{subsec:data_acquisition}).
Moreover, the dataset contains \textit{train} and \textit{test} camera trajectories recorded under the following four different lighting conditions:

\begin{itemize}
    \item \textit{natural} – Natural light of the environment
    \item \textit{cold} – Cold artificial light
    \item \textit{warm} – Warm artificial light
    \item \textit{dark} – No light.
\end{itemize}

\begin{figure}[t]
    \centering    
    \includegraphics[width=\linewidth]{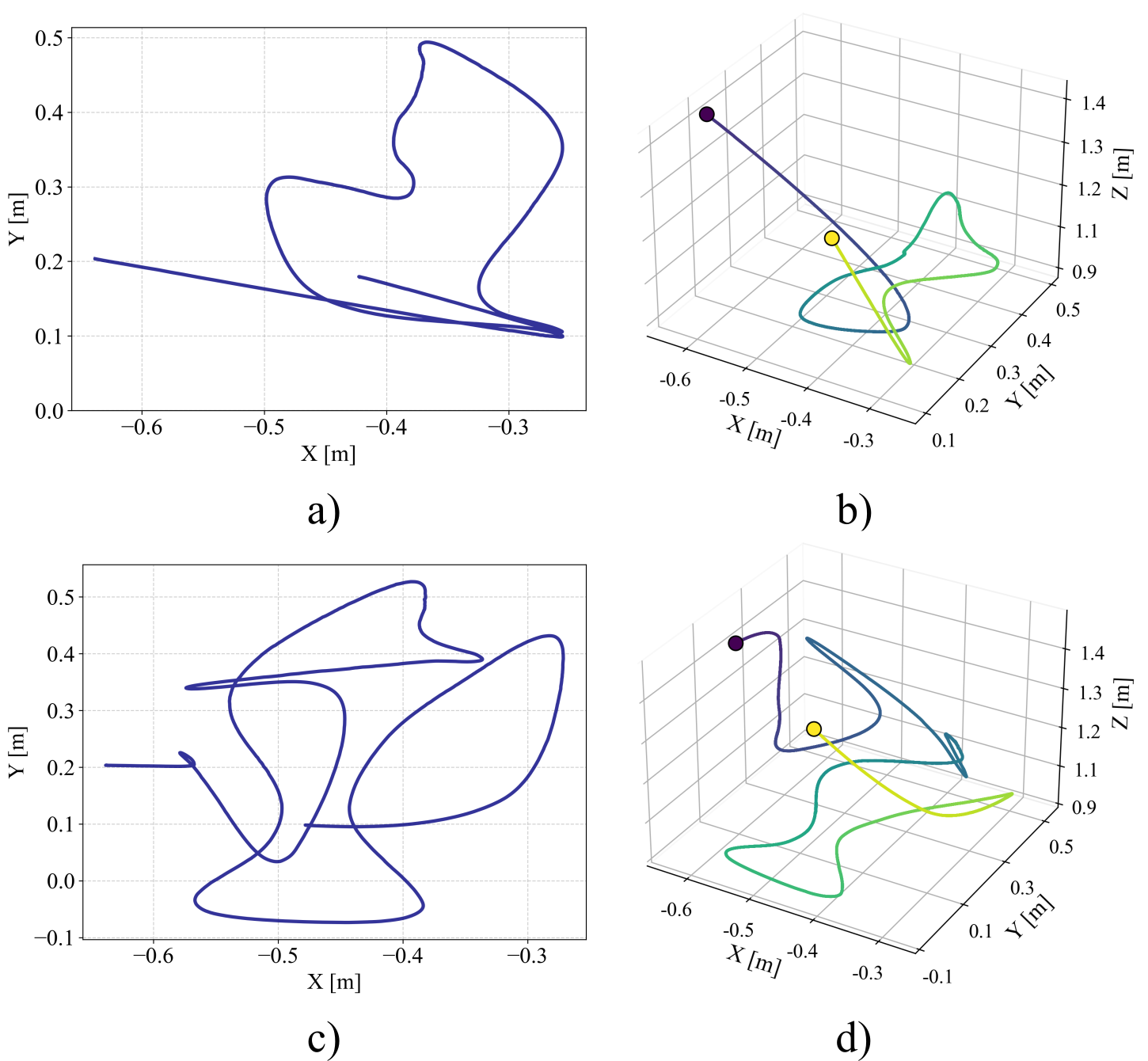}
    \caption{Our \ourdataset dataset contains train and test camera trajectories. From top left to bottom right: a) 2D representation of train trajectory, b) 3D representation of train trajectory, c) 2D representation of test trajectory, and d) 3D representation of test trajectory. Blue and yellow points represent start and end points of the trajectories, respectively.}
    \label{fig:trajectories}
\end{figure}

This enables our dataset to be used with classical SLAM approaches as well as newer Gaussian splatting- and NeRF-based paradigms, which are particularly sensitive to lighting conditions and may require proper training to perform effectively. Therefore, this dataset feature ensures that SLAM algorithms are tested and trained in environments that reflect real-world variability, ultimately evaluating their performance and reliability in practical applications. 
Although the KUKA LBR iisy 3 R760 manipulator has a repeatability parameter\footnote{The repeatability parameter measures the manipulator's error when returning to a previously reached position.} 
of $\pm\,0.1$\,mm, we release the joint configurations and flange poses for all the sequences belonging to train and test trajectories. As a result, the camera pose estimation conducted by any SLAM algorithm on this dataset remains unaffected by potential variations in joint configurations.
Furthermore, with smooth robot motions, the camera performs loop closures in both train and test trajectories (see \refFig{fig:trajectories}). {Train trajectories were designed to ensure full object visibility, while test trajectories intentionally include partial occlusions and viewpoint extrapolation to evaluate generalization.} 
The sequences include scenes with both opaque and transparent objects, reflective and non-reflective materials, as well as varying occlusions resulting from the rearrangement of objects.
The dataset features the following five distinct scenes (see \refFig{fig:objects_and_setups}):

\begin{itemize}
\item \textit{setup-1} and \textit{setup-2}: supermarket goods, including transparent objects;
\item \textit{setup-3}: supermarket goods with only opaque objects;
\item \textit{setup-4}: industrial objects;
\item \textit{setup-5}: industrial objects in a cluttered arrangement.
\end{itemize}

We additionally release chessboard images for intrinsic and extrinsic calibration. The calibration procedure and resulting parameters are detailed in \refSec{subsec:calib}.
Tools for data association, ROS2 bag data extraction, and synchronization can be found on our website.

\subsection{Data acquisition}
\label{subsec:data_acquisition}

The RGB-D images were recorded at full depth resolution 
(848×480) and full frame rate (30 Hz) with an Intel RealSense D435i camera. The IMU readings were acquired at a rate of 210 Hz. 
We used the library \texttt{Intel RealSense SDK 2.0}\footnote{\url{https://www.intelrealsense.com/sdk-2/}} 
to record the camera data. While the depth and color images are initially unaligned, as they are captured by two distinct sensors, we used the built-in feature of the Intel software library to register the depth images to the color images. 
The joint angles of the robot manipulator were collected 
at a frequency of 25 Hz.

To record ground-truth data, we employed an OptiTrack MCS consisting of twelve PrimeX 13W cameras installed around the robot cell. These cameras track the position and orientation of passive markers at 120 Hz.
The markers are mounted on a circular structure between the camera and the robot flange (see \refFig{fig:reference-frames}). 
This structure was printed using an industrial-grade 3D printer, ensuring high accuracy while tracking the markers. Additionally, we have added two markers on the flange and one marker on the camera itself to enhance the asymmetry of the marker setup.

To minimize reflections from the surrounding environment, non-reflective black panels were installed around the perimeter of the robot cell{, while a black cloth covered the robot support table.} Pictures featuring complex and colorful patterns have been placed on the black panels not to oversimplify the novel view synthesis of NeRF-based algorithms. 
This setup was chosen to ensure high-quality, accurate ground-truth data, as environmental reflections and lighting variations within the robot cell degraded the ground-truth data in preliminary trials.
The cold and warm lighting setups were obtained with a Neewer 660 (50W) PRO LED lamp.

\subsection{Dataset structure}
Inspired by \cite{TUMdataset}, our dataset is organized into sequences, which share the same structure across all setups. For the sake of simplicity, consider the following \textit{setup-1} structure:
\vspace{1.0mm}
{\fontsize{8}{9}\selectfont
    \dirtree{%
    .1 setup-1.
    .2 natural/.
    .3 \dots.
    .2 warm/.
    .3 \dots.
    .2 cold/.
    .3 \dots.
    .2 dark/.
    .3 \dots.
    .2 intrinsics.yaml.
    .2 extrinsics.yaml.
    }
}
\vspace{1.0mm}

Each lighting condition has a \textit{train} and \textit{test} sequence, which are organized equally. For example, the structure of the \textit{train} sequence is reported below.

\vspace{1.0mm}
{\fontsize{8}{9}\selectfont
    \dirtree{%
    .1 train.
    .2 rgb/.
    .3 1739371345.939813614.png.
    .3 1739371346.193981361.png.
    .3 \dots.
    .2 depth/.
    .3 1739371345.939813614.png.
    .3 1739371346.193981361.png.
    .3 \dots.
    .2 robot_data/.
    .3 joint_positions.txt.
    .3 flange_poses.txt.
    .2 associations.txt.
    .2 imu.txt.
    .2 groundtruth_raw.csv.
    .2 groundtruth.txt.
    }
}
\vspace{1.0mm}

These folders and files contain the following data:
\begin{itemize}
    \item \texttt{rgb/}: timestamped color images (\texttt{PNG} format, 3 channels, 8 bits per channel);
    \item \texttt{depth/}: timestamped {raw} depth images (PNG format, 1 channel, 16 bits per channel, distance scaled by a factor of 1000). The depth images are already aligned with the RGB sensor;
    \item \texttt{associations.txt}: list of all the temporal associations between the RGB and depth images (format: \texttt{rgb\_timestamp rgb\_file depth\_timestamp depth\_file});
    \item \texttt{imu.txt}: timestamped accelerometer and gyroscope readings (format: \texttt{timestamp ax ay az gx gy gz});
    \item \texttt{groundtruth.txt}: timestamped ground truth of the camera poses w.r.t world reference frame, expressed as a sequence of quaternions and translation vectors  (format: \texttt{timestamp qx qy qz qw tx ty tz}); 
    \item \texttt{groundtruth_raw.csv}: MCS ground-truth raw data of flange marker ring trajectory;
    \item \texttt{extrinsics.yaml}: transformation matrices of the reference frames involved in the data collection process (see \refSec{subsec:ref-frames});
    \item \texttt{instrinsics.yaml}: intrinsic calibration parameters of the camera; 
    \item \texttt{robot\_data/flange\_poses.txt}: timestamped poses of the robot flange. Each pose is expressed w.r.t. the robot base as a quaternion and translation vector (format: \texttt{timestamp qx qy qz qw tx ty tz});
    \item \texttt{robot\_data/joint_positions.txt}: timestamped angles of each robot joint (format: \texttt{timestamp q1 q2 q3 q4 q5 q6}).
\end{itemize}
The accelerometer data are provided in $\mathrm{m/s^2}$, the gyroscope data in $\mathrm{rad/s}$, and the ground truth and flange positions in $\mathrm{mm}$.
All sequences are also available as ROS2 bag files.

\subsection{Reference frames}
\label{subsec:ref-frames}
\begin{figure}[t]
    \centering    
    \includegraphics[scale=0.15]{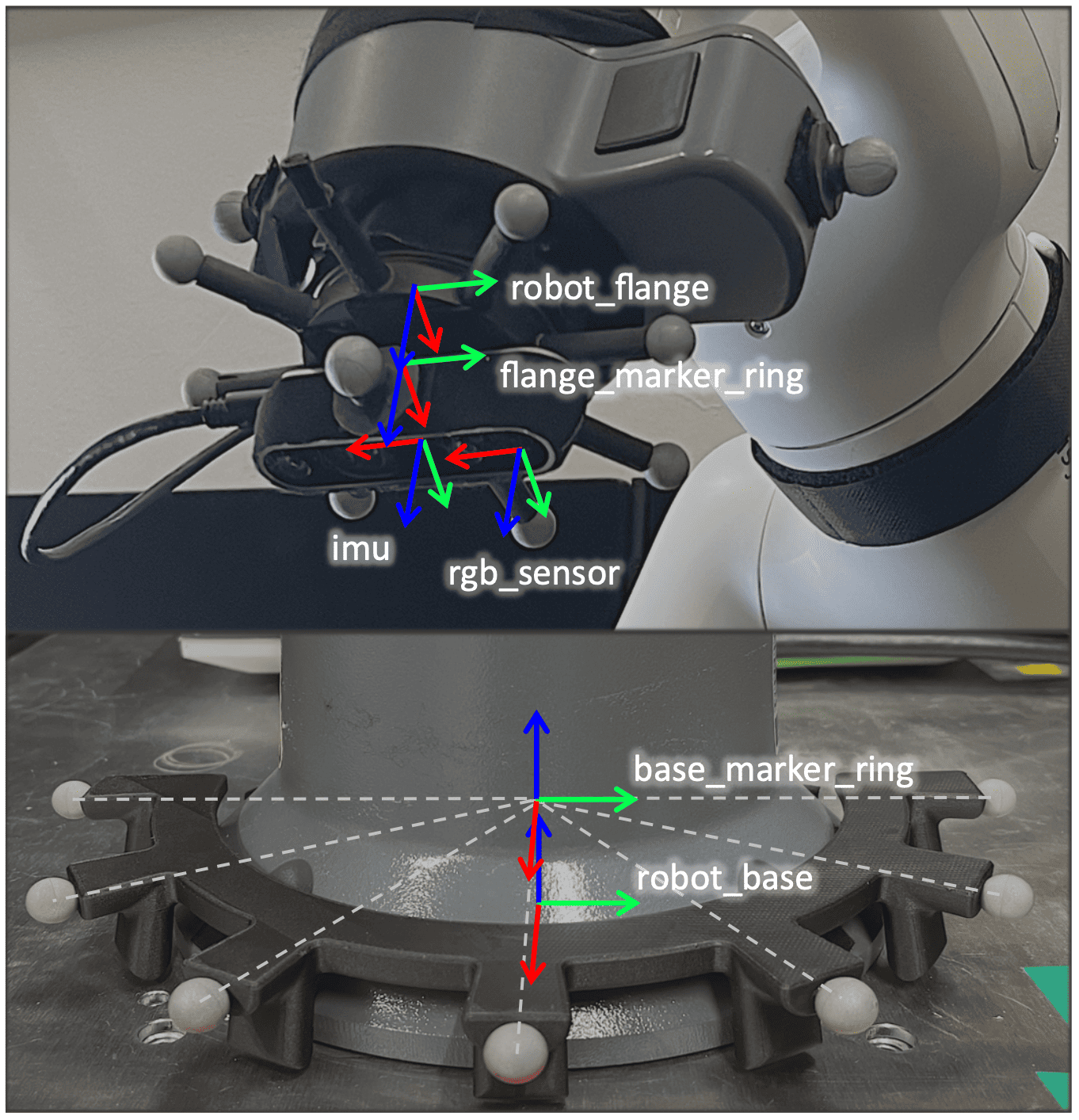}
    \caption{Illustration of the reference frames involved in the data collection.}
    \label{fig:reference-frames}
\end{figure}

The reference frames shown in \refFig{fig:reference-frames} were used in the data collection process to align the data coming from the camera, the robot and the MCS. These reference frames are described below.

\begin{itemize}
    \item \texttt{robot_flange}: defined at the center of the robot flange;
    \item \texttt{flange_marker_ring}: used by the MCS to locate the robot flange;
    \item \texttt{rgb_sensor}: optical center of the RGB sensor;
    \item \texttt{imu}: IMU sensor pose;
    \item \texttt{base_marker_ring}: used by the MCS to locate the robot base;
    \item \texttt{robot_base}: defined at the center of the robot base.
\end{itemize}
\subsection{Motion Capture System (MCS)}
\label{subsec:mocap}
\textbf{Calibration.}~To collect ground-truth data, we used twelve Optitrack PrimeX 13W cameras, carefully positioned to focus on the robot flange from various angles and minimize the occlusions caused by the robot motion.
The cameras were calibrated with the Motive software\footnote{\url{https://optitrack.com/software/motive/}} by Optitrack. 
The 3D mean error of the calibration was $0.4$\,mm, representing the convergence error between the twelve cameras.

\begin{figure}[b]
    \centering
    \includegraphics[scale=1.5]{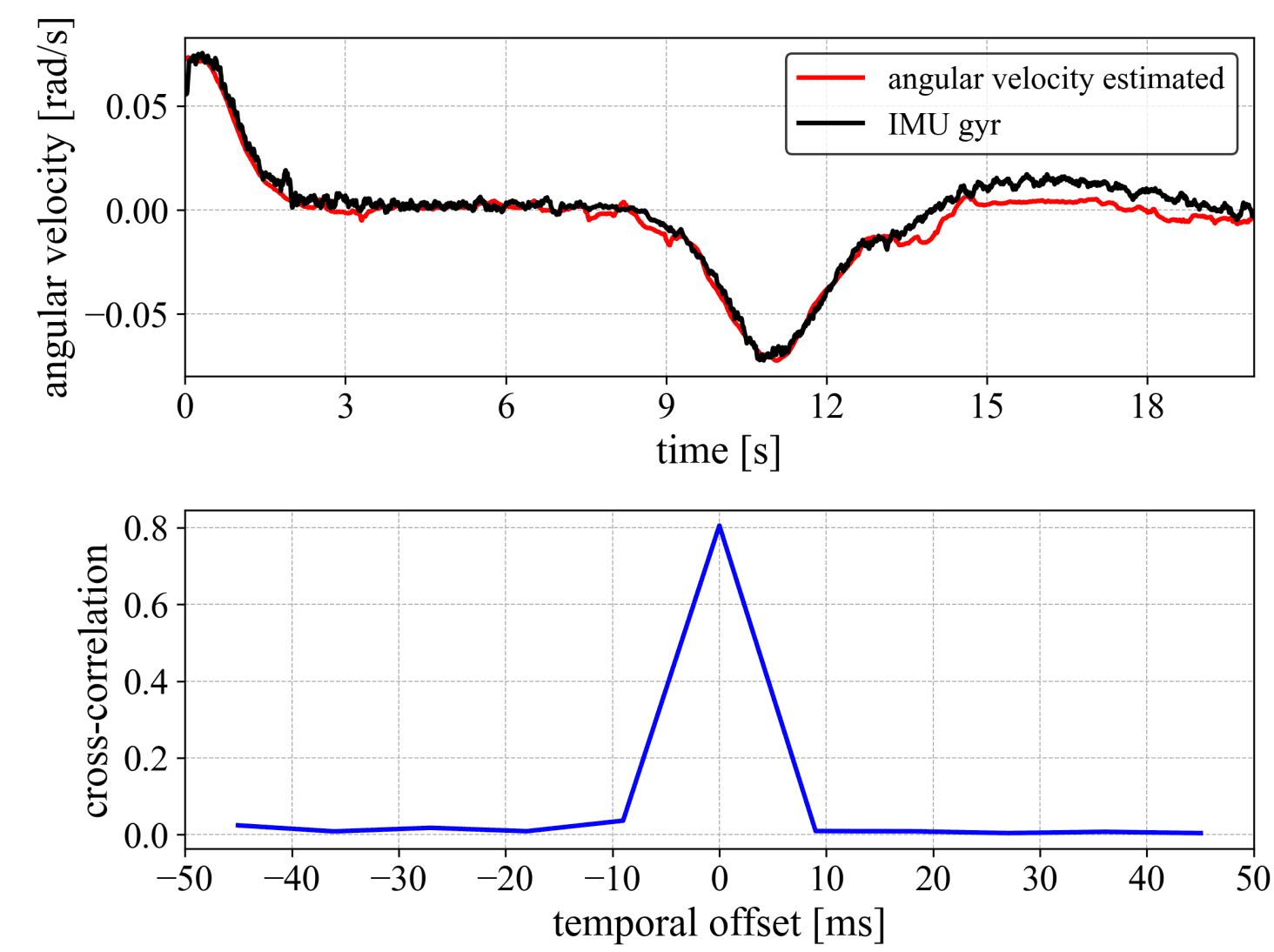}
    \caption{ Validation of time-synchronized data {by comparing angular velocity estimates from the MCS and the IMU gyroscope measurements.}}
    \label{fig:sync}
\end{figure}
\textbf{Synchronization.}~{All sensing modalities are temporally aligned using a unified Unix Epoch time reference. Hardware timestamps are preserved whenever available and subsequently refined via cross-correlation validation.}
Specifically, the data gathered from the MCS are initially unsynchronized.
To ensure temporal alignment, we retrieved for each sequence the \textit{capture starting time} provided by OptiTrack.
It was used to compute the Unix Epoch timestamp for each pose reported in the provided ground-truth file.
Following the approach presented in \cite{bonarini2006rawseeds}, the synchronization was validated by comparing the angular velocity estimated from the MCS with the gyroscope measurements from the IMU using cross-correlation (see \refFig{fig:sync}).
The maximum at zero offset confirms correct alignment.

\subsection{Camera calibration}
\label{subsec:calib}
\begin{figure}[t]
    \centering    
    \includegraphics[width=0.9\linewidth]{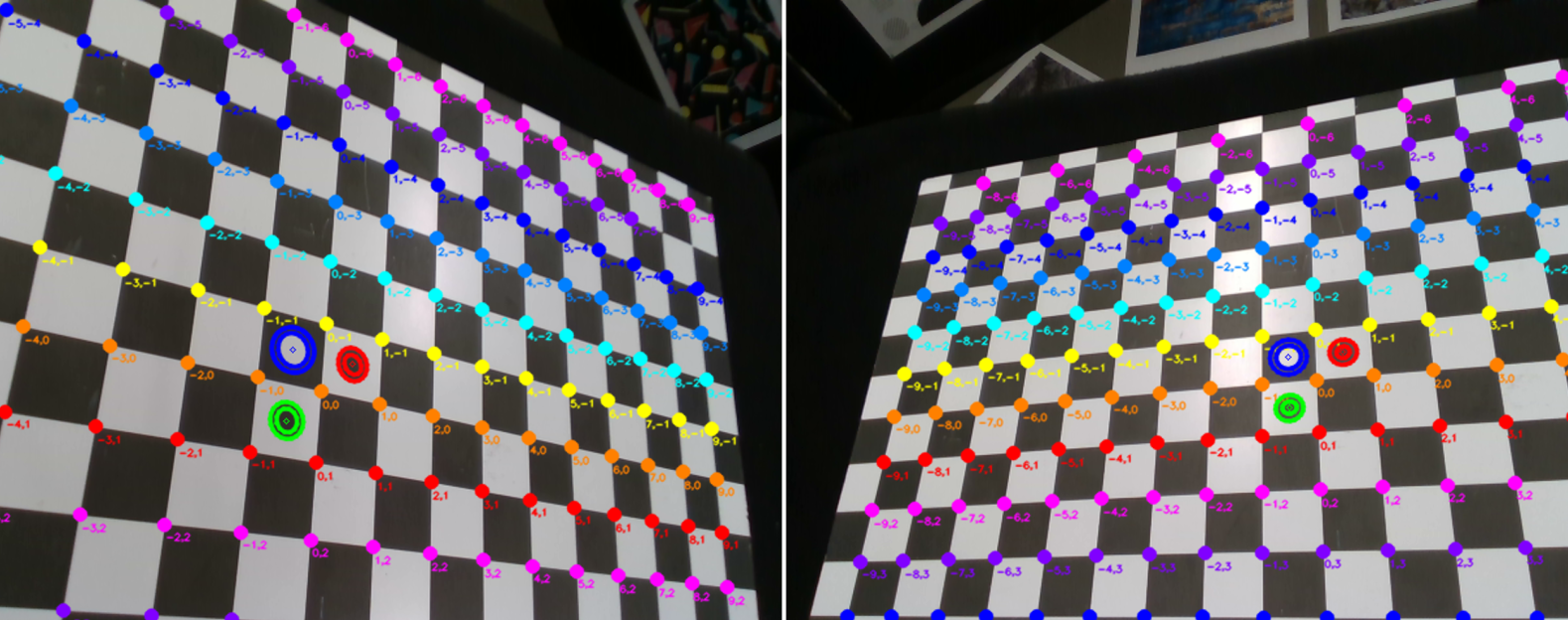}
    \caption{Corners and fiducial markers at the calibration procedure.}
    \label{fig:detected_patterns}
\end{figure}

To estimate the hand-eye transformation (i.e., the transformation from the robot flange to the camera projection plane), more than 200 images containing the calibration plate were captured, and the corresponding robot pose for each image was saved. Subsequently, the calibration method presented in \cite{shiu1989} was applied. The calibration plate (a chessboard with squares of $30 \times 30~\mathrm{mm}$) was detected by applying a custom detection method (see Fig. \ref{fig:detected_patterns}). 
For a more accurate calibration, not only the extrinsic parameters (hand-eye transformation) were estimated, but also the intrinsic parameters.
Our experiments have shown that the use of the estimated intrinsic parameters resulted in more accurate extrinsic parameters.
The intrinsic parameters of the camera (focal length (fx/fy), principal point (cx/cy) and camera distortion coefficients) are shown in \refTab{tab:intrinsics-values}. The extrinsic parameters are shown in the \refTab{tab:extrinsics-values}.
\begin{table}[t]
\footnotesize
\centering
\setlength{\tabcolsep}{4.5pt} 
\rowcolors[]{2}{white} {gray!15}
\resizebox{0.42\textwidth}{!}{
\begin{tabular}{l c c c c c c c c c}
    {Parameters} & {fx}~[$\mathrm{px}$] & {fy}~[$\mathrm{px}$] & {cx}~[$\mathrm{px}$] & {cy}~[$\mathrm{px}$] & {distortion coefficients} \\
    \toprule
    factory  & 605.2 & 605.1 & 425.7 & 246.0 & 0 0 0 0 0  \\
    estimated  & 599.6 & 599.2 & 426.6 & 246.3 & 0.1 -0.4 0 0 0.3\\
    \bottomrule
\end{tabular}
}
\caption{{Intrinsic camera parameters used in our dataset.}}
\label{tab:intrinsics-values}
\end{table}
\begin{table}[t]
\footnotesize
\centering
\setlength{\tabcolsep}{5pt} 
\rowcolors[]{2}{white} {gray!15}
\resizebox{0.485\textwidth}{!}{
\begin{tabular}{l c c c c c c c}
    {Used intrinsics} & {tx} & {ty} & {tz} & {qx} & {qy} & {qz} & {qw} \\
    \toprule
    factory   & 0.5895  & 32.5419 & 46.0707 & 0.0060  & 0.0041  & -0.7035 & 0.7106 \\
    estimated & -0.2852 & 32.7894 & 46.1513 & 0.0063  & 0.0054  & -0.7037 & 0.7104 \\
    \bottomrule
\end{tabular}
}
\caption{{Extrinsic parameters (hand–eye transformation): translation [$\mathrm{mm}$] and unit quaternion rotation.}}
\label{tab:extrinsics-values}
\end{table}

\begin{table}[t]
\footnotesize
\centering
\setlength{\tabcolsep}{1.5pt}
\rowcolors[]{2}{white} {gray!15}
\resizebox{0.486\textwidth}{!}{
\begin{tabular}{l c ccc}
     & {Reproj. error} &\multicolumn{3}{c}{{Hand-eye}}\\
    \cmidrule(lr){3-5}
    {} & {[$\mathrm{px}$]} & {Trans.\,[$\mathrm{mm}$]} & {Rot.\,[$\mathrm{deg.}$]} & {Rel. transf. err.\,[$\mathrm{a.u.}$]}\\
    \toprule
    Mean & 0.36 & 2.38 & 0.33 & 2.90 \\
    Std. dev. & 0.04 & 0.41 & 0.08 & 0.66 \\
    Std. error & 0.01 & 0.13 & 0.02 & 0.21 \\
    \midrule
    \textbf{All images (estim. intrinsics)} & \textbf{0.38} & \textbf{2.31} & \textbf{0.31} & \textbf{2.69} \\
    All images (factory intrinsics) & 0.61 & 2.37 & 0.33 & 2.77 \\
    \bottomrule
\end{tabular}
}
\caption{Mean, standard deviation, and standard error of calibration metrics. The last two rows use all images with estimated vs. factory intrinsics.}
\label{tab:calib-metrics}
\end{table}

To compare the accuracy of the calibration using both factory and estimated intrinsic parameters, the set of calibration images was divided into several batches (10 batches with 20 images in each), and the same calibration process was applied to each batch. The accuracy of the calibration was assessed using the metrics proposed in \cite{10.1371/journal.pone.0273261}, which include translation error in millimeters (\textit{Trans.}\,[mm]), rotation error in degrees (\textit{Rot.}\,[deg.]), and unitless relative transformation error (\textit{Rel. transf. error}\,[a.u.]). \refTab{tab:calib-metrics} shows the mean, standard deviation, and standard error of the calibration metrics. The accuracy of the hand-eye transformation, calculated using the estimated intrinsic parameters, showed more precise results, with reduced reprojection, translation, rotation, and relative transformation errors. 

\section{Experiments}
\label{experiments}
{To evaluate the dataset's suitability for benchmarking methods at the intersection of SLAM and Novel View Synthesis, we present results from state-of-the-art techniques.}

\subsection{Metrics}
We assess camera tracking performance using the Root Mean Square Error of the Absolute Trajectory Error (ATE RMSE~$\downarrow$). 
Following~\cite{kerbl20233d}, to evaluate NVS, we report Peak Signal-to-Noise Ratio (PSNR~$\uparrow$), Structural Similarity Index Measure (SSIM~$\uparrow$), and Learned Perceptual Image Patch Similarity (LPIPS~$\downarrow$). PSNR captures pixel-level differences, while SSIM reflects perceptual similarity in structure, luminance, and contrast~\cite{ssim}. Finally, LPIPS measures perceptual fidelity using deep feature distances~\cite{Zhang_2018_CVPR}.


\subsection{Results}


\textbf{Novel View Synthesis (NVS).} 
To highlight the value of independent test trajectories, we evaluated two state-of-the-art baselines: Gaussian Splatting (GSPLAT)~\cite{kerbl20233d} and FeatSplat (FSPLAT)~\cite{martins2024}, the latter designed to mitigate viewpoint overfitting. 
In this experiment, \textit{setup-1}, \textit{setup-3}, and \textit{setup-4} were used under natural, warm, and dark lighting conditions.
Baselines were initialized with random point clouds and a black background: GSPLAT used four spherical-harmonics bands, and FSPLAT used 32-dimensional latent vectors and a multilayer perceptron with one hidden layer of 64 units.
They were optimized for a total of 21k iterations, and we report metrics both at 7k and 21k iterations.
The baselines were optimized on the respective \textit{train} trajectories of each setup, skipping one every ten frames.
We evaluate both with the common approach of using skipped frames from the \textit{train} trajectory~\cite{mildenhall2021nerf, kerbl20233d} (\refTab{tab:nvs_classic}) and using the independently recorded \textit{test} trajectory of each setup (\refTab{tab:nvs_independent}).

\begin{table}[b] 
\footnotesize
\setlength{\tabcolsep}{4pt}
\centering
\resizebox{0.485\textwidth}{!}{
\rowcolors[]{2}{white} {gray!15}
\begin{tabular}{l  ccc | ccc | ccc}
   & \multicolumn{3}{c}{natural} & \multicolumn{3}{c}{warm} & \multicolumn{3}{c}{dark} \\
   \cmidrule(lr){2-4} \cmidrule(lr){5-7} \cmidrule(lr){8-10} 
    & PSNR & LPIPS & SSIM & PSNR & LPIPS & SSIM & PSNR & LPIPS & SSIM \\
   \midrule
   gsplat-7k & 17.32 & 0.445 & 0.694 & 17.67 & 0.497 & 0.609 & 23.77 & 0.464 & 0.678 \\
   fsplat-7k & 17.40 & 0.445 & 0.700 & 17.75 & 0.499 & 0.610 & 23.43 & 0.467 & 0.687 \\
   \hline
   gsplat-21k & 18.43 & 0.411 & 0.724 & 18.75 & 0.468 & 0.637 & 24.99 & 0.443 & 0.708 \\
   \textbf{fsplat-21k} & \textbf{19.45} & \textbf{0.406} & \textbf{0.743} & \textbf{20.01} & \textbf{0.461} & \textbf{0.656} & \textbf{25.13} & \textbf{0.444} & \textbf{0.714} \\
   \bottomrule
\end{tabular}
}
\caption{Classic evaluation of Gaussian Splatting (gsplat) and FeatSplat (fsplat) reporting PSNR[$\uparrow$], LPIPS[$\downarrow$], and SSIM[$\uparrow$].}
\label{tab:nvs_classic}
\end{table}

\begin{figure}[b]
    \centering
    \includegraphics[width=0.99\linewidth]{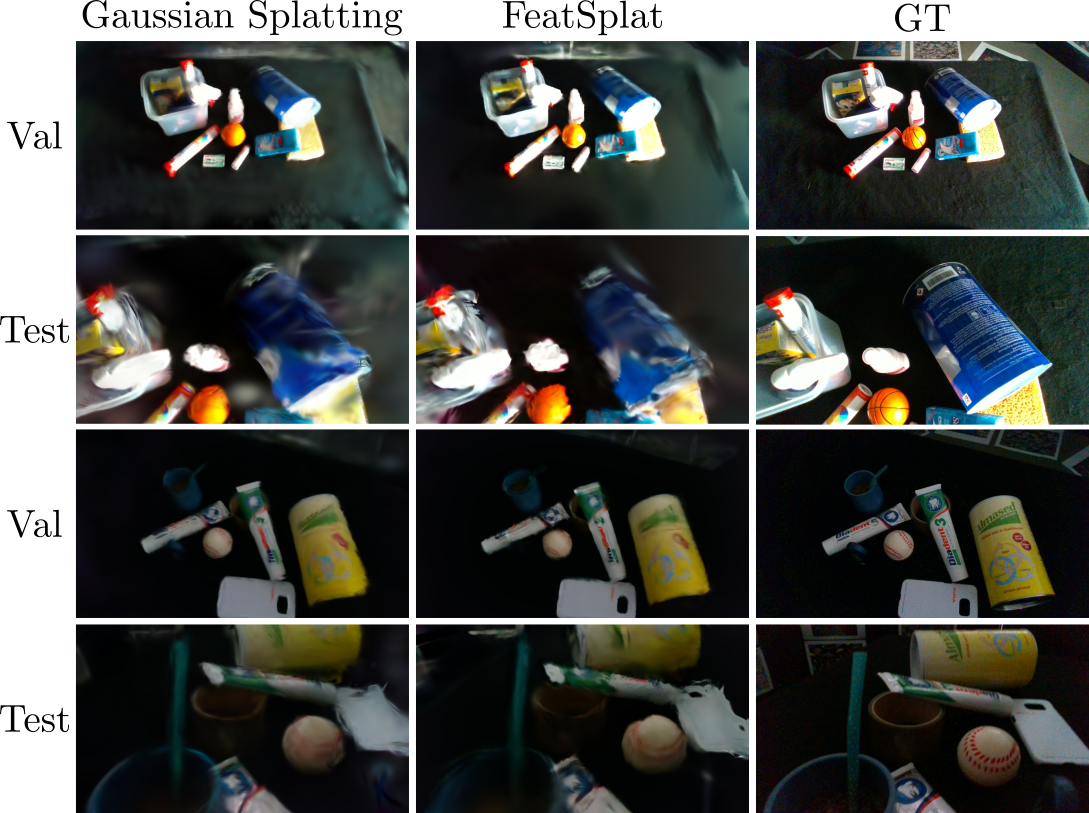}    
    \caption{Comparison of NVS for validation and test viewpoints. Although both models perform well when evaluated on validation viewpoints of the training trajectory, the independent test trajectory reveals overfitting.}
    \label{fig:nvs_val_vs_test}
\end{figure}

\begin{table}[t] 
\footnotesize
\centering
\setlength{\tabcolsep}{4pt} 
\resizebox{0.485\textwidth}{!}{
\rowcolors[]{2}{white} {gray!15}
\begin{tabular}{l  ccc | ccc | ccc}
   & \multicolumn{3}{c}{natural} & \multicolumn{3}{c}{warm} & \multicolumn{3}{c}{dark} \\
   \cmidrule(lr){2-4} \cmidrule(lr){5-7} \cmidrule(lr){8-10} 
    & PSNR & LPIPS & SSIM & PSNR & LPIPS & SSIM & PSNR & LPIPS & SSIM \\
   \midrule
   gsplat-7k & 14.53 & 0.493 & 0.608 & \textbf{15.45} & 0.543 & 0.527 & 21.82 & 0.499 & \textbf{0.596} \\
   fsplat-7k & \textbf{14.73} & 0.488 &\textbf{ 0.623 }& 15.42 & 0.541 & \textbf{0.540} & 21.73 & 0.498 & 0.630 \\
   \hline
   gsplat-21k & 14.42 & \textbf{0.478} & 0.606 & 15.35 & \textbf{0.531} & 0.524 & 21.53 & 0.492 & 0.600 \\
   fsplat-21k & 14.71 & \textbf{0.478} & 0.614 & 15.29 & 0.532 & 0.535 & \textbf{21.91} & \textbf{0.490} & 0.627 \\
   \bottomrule
\end{tabular}
}
\caption{Evaluation on independent test trajectory of Gaussian Splatting (gsplat) and FeatSplat (fsplat)  reporting PSNR[$\uparrow$], LPIPS[$\downarrow$], and SSIM[$\uparrow$].}
\label{tab:nvs_independent}
\end{table}%
The common evaluation indicates an improvement in rendering performance, for both baselines, through the complete optimization process (see \refTab{tab:nvs_classic}). 
Both baselines achieve their best metrics under \textit{dark} lighting conditions. {This is probably due to using black as initial color for the Gaussians and while rendering artifacts are less visible it does not imply a better reconstruction of the 3D structure.} 
Nevertheless, the test trajectory at \ourdataset reveals that both baselines overfit to the training trajectory. Note how the rendering quality does not improve beyond the 7k iteration and even slightly declines in \refTab{tab:nvs_independent}.
Qualitative analysis in \refFig{fig:nvs_val_vs_test} also showcases the contrast between the high-quality rendering of validation viewpoints and the {floating artifacts generated to fit training views that a clearly visible} in test viewpoints.
This result validates the need for datasets with independent trajectories to evaluate novel view synthesis methods and shows that the classic approach of sampling evaluation frames from the training trajectory does not inform about trajectory overfitting.

\begin{table}[t]
\footnotesize
\centering
\setlength{\tabcolsep}{6pt}
\resizebox{0.45\textwidth}{!}{
\rowcolors[]{2}{white} {gray!15}
\begin{tabular}{l l l | c c c}
    {Setup} & {Ill.} & {Traj.} & {MonoGS} & {Kinematics} & {MonoGS (Kinem. seed)}\\
    \toprule
    setup-1 & natural & train & 3.3 & 0.5 & 2.0 \\
      & natural & test  & 3.9 & 0.4 & 2.1 \\
      & cold & test  & 2.4 & 0.3 & 2.2 \\   
      & warm & test  & 2.1 & 0.3 & 1.4 \\
      & dark & test  & 3.1 & 0.4 & 1.9 \\
    \midrule
    setup-3 & natural & train & 2.6 & 0.5 & 1.1 \\
      & natural & test  & 3.7 & 0.4 & 1.7 \\
      & dark & train  & 0.6  & 0.5  & 0.6 \\
      & dark & test  & 3.8  & 0.3 &  1.6 \\
    \midrule
    setup-4 & natural & train & 0.9 & 0.5 & 0.7 \\
     & cold    & train & 2.4 & 0.5 & 1.0 \\
     & warm    & train & 1.6 & 0.5 & 0.8 \\
     & dark    & train & 1.0 & 0.5 & 0.9 \\
    \bottomrule
\end{tabular}
}
\caption{ATE RMSE [$\downarrow \mathrm{cm}$] across multiple sequences using three algorithm configurations: MonoGS, Kinematics and MonoGS with Kinematic data as seed.}
\label{tab:ate-comparison}
\end{table}

\textbf{Camera tracking. }
To illustrate the advantage on using kinematic data, three approaches based on Gaussian Splatting SLAM \cite{matsuki2024gaussian} (MonoGS) were evaluated.
In the first experiment, we used the default configuration for MonoGS, while in the second one we integrated the robot kinematic data to describe the camera motion. 
The evaluation of the second approach is referred to as the \textit{Kinematics} experiment. 

\begin{figure}[b]
    \centering
    \includegraphics[width=0.99\linewidth]{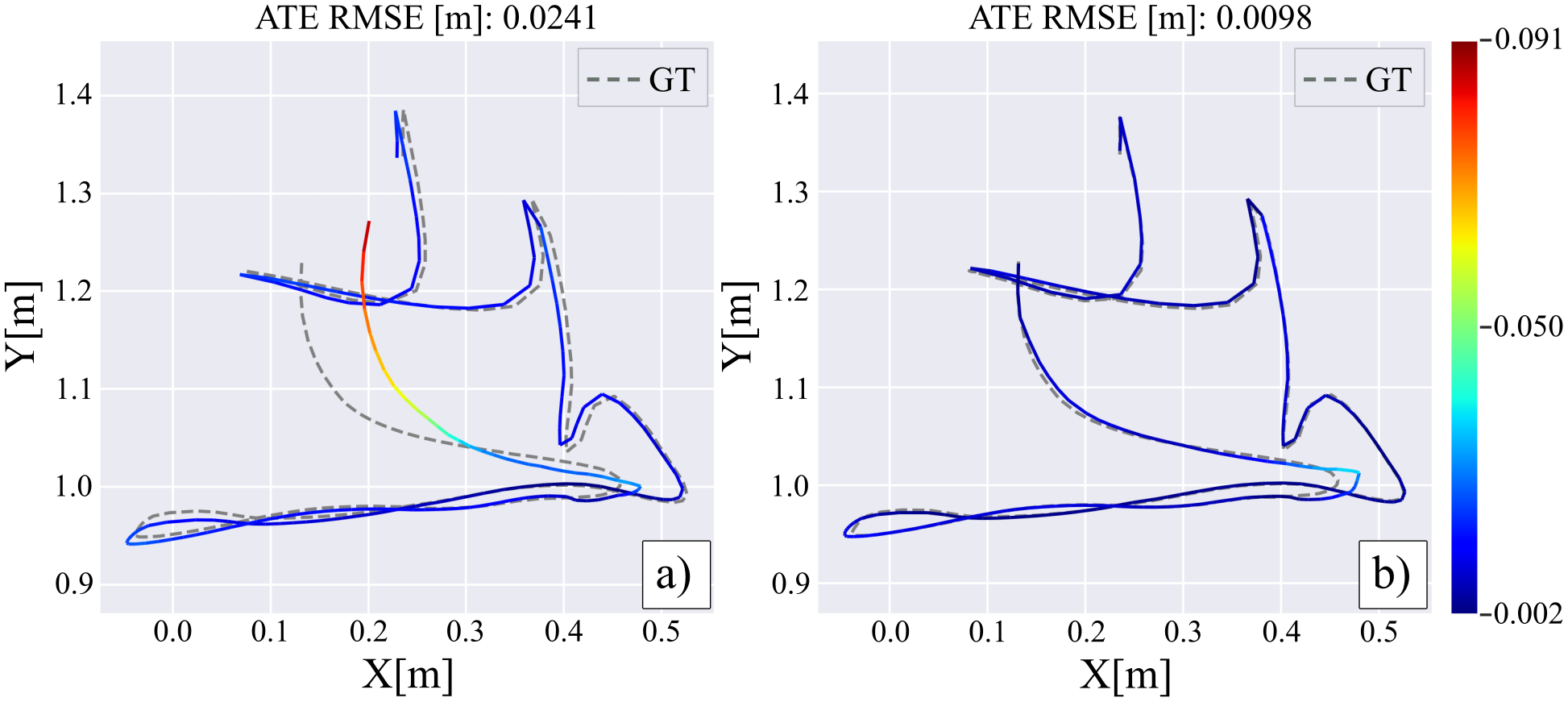}
    \caption{Camera tracking result for \textit{setup-4} under cold light using a) MonoGS and b) MonoGS with kinematic data as initial seed. Dashed line represent the ground-truth while colored line represent the camera trajectory. Incorporating kinematic data allows drift correction in camera tracking.}
\label{fig:ate-trajectories}
\end{figure}

In the third experiment, robot kinematics were incorporated as an initial seed for camera pose estimation within MonoGS. This is referred to as \textit{Kinem. seed} experiment.
Following the evaluation protocol in \cite{matsuki2024gaussian}, the ATE RMSE for keyframes is reported in \refTab{tab:ate-comparison}. For this experiment, \textit{setup-1}, \textit{setup-3}, and \textit{setup-4} were used under different light conditions while performing \textit{train} and \textit{test} trajectories.
In all cases, it is evident that incorporating kinematic data improves camera tracking performance, allowing for drift corrections. However, the results also show that the naive approach of using robot kinematics as the initial seed of MonoGS decreases its accuracy.
{Differently from continuous integration, which introduces corrective constraints throughout the optimization process, using the kinematics solely for initialization can lead to reduced performance, as early pose biases tend to propagate into the subsequent photometric optimization.} 
This result motivates the use of our \ourdataset dataset for research on multimodal fusion within NeRF and Gaussian Splatting SLAM.
\refFig{fig:ate-trajectories} illustrates the camera tracking results for \textit{setup-4} under cold lighting condition. 
The trajectory using MonoGS is shown in \refFig{fig:ate-trajectories}a, while \refFig{fig:ate-trajectories}b shows the result using kinematic data as initial seed. 

\section{Conclusions}
\noindent
In this work, we introduced \ourdataset, a dataset for benchmarking methods that combine SLAM with Neural Rendering and Gaussian Splatting. Our dataset includes 40 sequences of synchronized RGB-D images, IMU readings, robot kinematic data, intrinsic and extrinsic camera parameters, and ground-truth pose data across diverse scenes and conditions. 
\ourdataset addresses research challenges such as generalization across reproducible trajectories and robustness to scene and illumination changes. By releasing robot kinematic data, it  also enables the assessment of novel SLAM strategies when applied to robot manipulators. It aims to drive future research in learning-based and hybrid methods requiring accurate localization and multi-sensor data fusion, ultimately enhancing the deployment of systems that combine perception, mapping, and rendering.
{Future extensions include larger workspaces, dynamic scenes (possibly with motion blur and abrupt motions), wider range of objects, and integration with mobile platforms.}
These enhancements will strengthen robustness and generalization tests, while maintaining the dataset's core focus.





\end{document}